# Spine Landmark Localization with combining of Heatmap Regression and Direct Coordinate Regression

Wanhong Huang, DLUT
Chunxi Yang, DLUT
Tianhong Hou, DLUT

## Abstract

*Landmark Localization plays a very important role in processing medical images as well as in disease identification. However, In medical field, it's a challenging task because of the complexity of medical images and the high requirement of accuracy for disease identification and treatment.*

*There are two dominant ways to regress landmark coordination, one using the full convolutional network (FCN* [6]*) to regress the heatmaps of landmarks (Commonly, it's using Gaussian heatmap), which is a complex way and heatmap post-process strategies are needed, and the other way is to regress the coordination using CNN + Full Connective Network (CNN + FC) directly, which is very simple and faster training , but larger dataset and deeper model are needed to achieve higher accuracy. Though with data augmentation and deeper network it can reach a reasonable accuracy, but the accuracy still not reach the requirement of medical field. In addition, a deeper networks also means larger space consumption.*

*Our contributions are: 1) To achieve a higher accuracy, we contrived a new landmark regression method which combing heatmap regression*[3]  *and direct coordinate regression base on probability methods and system control theory. And it improve 39.1% accuracy with compare to only using directly regression model. And achieve a accuracy of 99.6% in validation dataset. 2) We compare different ways of heatmap regression model and chosen one is reasonable in time and space's consumption. And We also compare the different of heatmap regression and directly coordinates regression.*

1. Introduction

landmark localization occupies a very important place in medical AI as It can help locate various complex tissue and structures and apply them to subsequent tasks such as disease diagnosis and treatment resolution generation.

However, landmark localization is a challenging task for medical image. There are several challenging points: 1) Medical AI require high precision of landmark coordination because these coordinates will be used for subsequent tasks such as disease diagnosis and treatment resolution generation. If we can't assure the accuracy of landmark localization, it would greatly increase the error of subsequent tasks. And these errors are fatal in medical industry.  So when landmark localization comes to  medical images, the accuracy become more crucial.  2) labeled medical images is rare, a dataset usually has only hundreds of labeled images. In our experiment ,what we use it a dataset with only 150 labeled images. We need to devise a model to take full advantages of these images. Though we can use data augmentation to enlarge our dataset, but it's effect is also limited. 3) Medical images are complex and diverse. Unlike common semantic segmentation or object recognition, which the same object has a very similar appearance. In medical images, even the same tissue or organ may have different appearance, and different persons' organ may also different. This makes it more difficult to locate the key points in medical images. In our experiment we would only use the spine's MRI (Magnetic Resonance Imaging) images.

Usually, we have two ways to complete a coordinate regression task. The first way is using CNN + FC. Which is a very simple way. It a end-to-end model. What we need is just put a Full Connective Layer after the CNN.

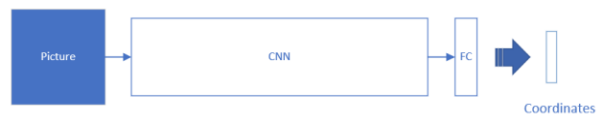

Figure 1. Using CNN + FC predict the coordinates directly

However, In our experiment, we superisingly found that CNN+FC performce well under a augumented dataset. It can predicted 62% points successfully (under a limit of 8mm bias).We found that though CNN + FC is a simple approach for landmark localization task, it may be also a feasible resolution if the dataset is large enough and the CNN network is deep enough.

The othe way commonly used to solve the landmark localization problem is heatmap regression. It pre-generate a heatmap base on the labeled landmark position. And then make CNN to predict the heatmap.

Because no FC layer participate in the trainning process. it can take advantage of CNN as it can preserved more spatial information.

Heatmap Regression [3] is a concept widely used in landmark localization and Semantic segmentation task. The target labels are heatmaps (Commonly it's Gaussian Heatmap) . In Figure 2. It shows a spine image and a heatmap predicted by network, which still not perfectly fit the spine data and a predicted heatmap that fit the data well.

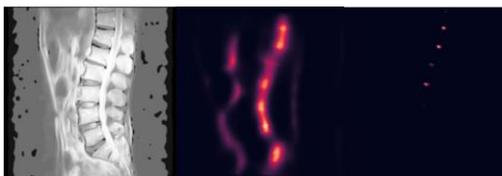

Figure 2. A spine's image and two heatmap generated from it

Different from direct coordinate regression. The heatmap describe a probability distribution rather than a specific coordinate or value. We can use Argmax to get the coordinate directly, but usually it's imprecise. To acquire a accurate value, we need some strategies to postprocess the heatmap. With compare to direct coordinates regression. It's more difficult to use heatmap regression, because it's not a end-to-end model, and it cost more time and space in the train process. However, it can achieve a relatively high accuracy with training in a small dataset. On the contrary, CNN + FC need more large dataset and deeper network for it to reach a high accuracy. But it can training faster and takes less space consumption.

As you can see in the Figure 3. The predicted heatmaps often contains multiple Gaussian distribution points. If we use argmax directly to generate the predicted coordinates, sometimes it will lead to predict to the adjacent point, which will lead to a large coordinate deviation.

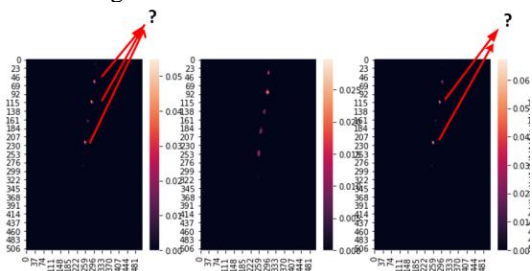

Figure 3. Predicted heatmaps with ambiguity

In fact, heatmap regression is relatively accurate in predicting the peak position of a single Gaussian distribution, but often multiple Gaussian distributions are predicted in one heatmap, which will lead to the prediction of adjacent point, resulting in large deviation (especially in the task of spinal landmark regression, the adjacent points are very close).

In our experiment, we found that both heatmap regression and direct coordinates regression can reach a relatively high accuracy. The can predict 60%~70%'s landmark right. The CNN + FC has the disadvantage that over fitting. And the heatmap regression have the disadvantage that post-process strategies are complex. Though we can use apply Argmax directly in the predicted heatmaps , it can't achieve a high accuracy. Therefore, we propose to combine CNN + FC with heatmap regression. It can improve the result's accuracy in an easy way.

2. Relate Works

2.1. Full Convolution Network (FCN)

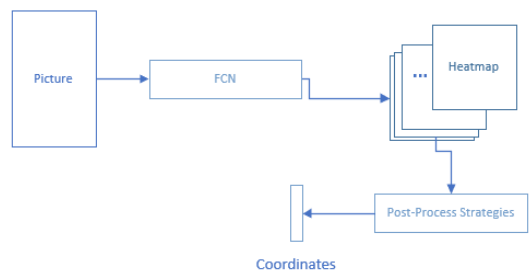

Figure 2. A Framework of landmark localization by FCN

FCN [6] is a kind of network which can realize image-to-image mapping. In order to implement the concept of heatmap regression, FCN. FCN uses up-sampling technique to map the reduced image features after convolution and pooling operations back to the image features with larger shapes (which can be regarded as the final feature heat map)

2.2. U-Net

U-Net ([12],[13]) is also a kind of network structure that can also realize the image-to-image mapping. It can be seemed as an improvement of FCN. And it has shown a high performance in many medical image semantic segmentation tasks. It uses a symmetrical U-shaped structure and a skip structure to combine different resolution information, because it uses the underlying features (cascade with the same resolution) to improve the lack of information in the up sampling. In addition, Medical image dataset are generally small, and the underlying features are very important.

2.3. Xcpetion

Xception [10] is an improvement of inception-v3, which has achieved good performance in the field of image classification. It has more than 24 million parameters. The structure can fit the data better with less parameters. Even compared with some networks, its efficiency is not very high, but in the case of high accuracy requirements, it is a good choice. Some research try to use Xception + FC to locate the landmarks directly. Although it has achieved a certain accuracy, it can not achieve the applicable level accuracy.

## 2.4. Spatial Configuration-Net

In the semantic segmentation of spine CT image, the author of [12] thinks that the common CNN + FC method needs a lot of data to ensure the accuracy. So they use the method based on regression heatmap to achieve semantic segmentation. In order to eliminate the influence of multiple Gaussian distribution points on the heatmaps, the heatmaps are put into SCN to continue training, and SCN also outputs some Gauss heatmaps. Finally, accurate heatmaps can be obtained by combining the heat maps from the output of the two networks.

## 3. Our Approaches

As mentioned earlier, the heatmaps output from the regression heatmap network often contains the peak points of Gaussian distribution, and the positions of the peak points of the Gaussian distribution are often more accurate. However, if the coordinates are obtained directly from these heatmaps, they are easy to be located to the adjacent points, which will cause large errors. Some post-processing strategies are needed to use for obtaining more accurate Gaussian heatmaps, such as SPN. On the contrary, CNN + FC approach can predict the position near the key point (the accuracy depends on the model.), and the error is not too large. In our experiments, we considered combining the two approaches.

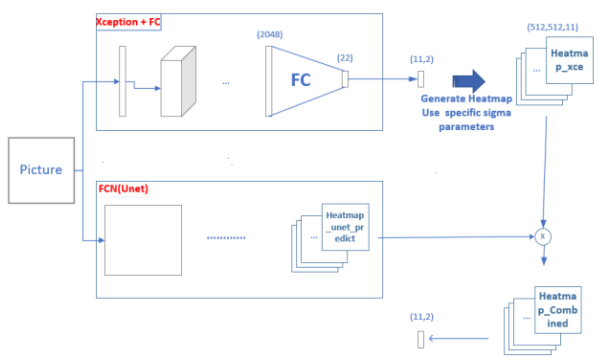

Figure 4. A simple structure of our approach

A simple structure of our scheme is shown in Figure 4. First, we use direct coordinate regression to train Xception + FC model. Then we use U-net model to regress Gauss heat maps. When both networks converge, for an input image, two models are used to predict the results. Xception will predict a coordinate result with shape of (11,2) and U-net will predict a result with shape of (512512,11) which represent heatmaps. After that, the coordinate results of Xception are used to generate heatmaps (the sigma parameters of Gaussian distribution can be specified according to experience or network training which shown in Figure 5. ), and finally the two are combined by product. Then we can use Argmax to get the accurate coordinates directly.

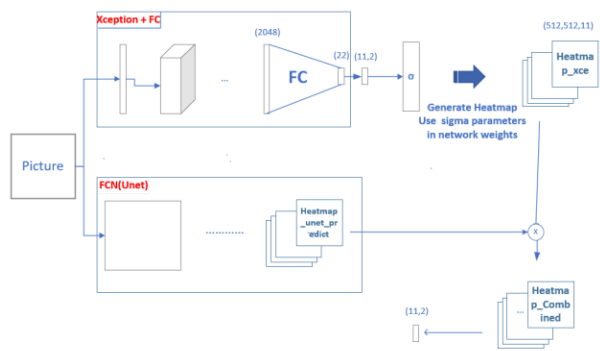

Figure 5. A structure of our approach with sigma parameters is in a network's weights

In our validation, this way reach a accuracy of 99.6%, with total of 550 landmarks and 548 landmark are located right. (Under the bias limit of 8mm)

### 3.1. Dataset Augmentation

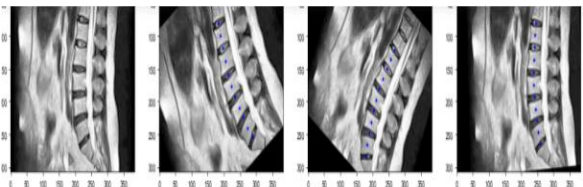

Figure 6. A augmented image

In order to make the models converged better and alleviate the over fitting phenomenon ( especially the direct coordinate regression model will have serious over fitting phenomenon in the case of small dataset ), we need to augment the dataset. And it should be noted that the augmentation of regression heat map training data set will consume a lot of space, so we use OpenCV augment the dataset dynamically.

We use translation, rotation and scale operations to augment an image. We use random translation of - 8 ~ 8 Pt on the y-axis, and -35 ~ 35 Pt on the x-axis. Because the spine is vertical, excessive vertical translation can cause information loss. And we rotate the image randomly in the angular range of -25° ~ 25°. The scale range is from 0.7 ~ 1.3.

### 3.2. Data Pre-Processing

Because some images are too dark, and there is a problem of low image contrast. In order to deal with these problems and make the model converge better, we use histogram equalization technique for each image.

227

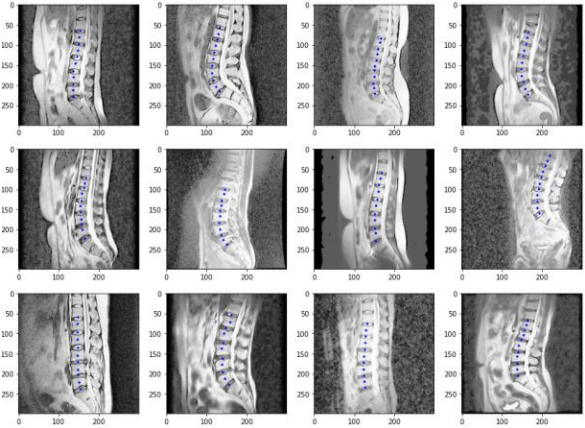

Figure 6. The dataset after pre-processing

And in order to make the image can be put into the network for training, we unify the size of the images. Use the size of (512,512) for U-net and (299,299) for Xception + FC. Then the labels need to be normalized, because they correspond to the horizontal and vertical coordinates of the original images, so we need to normalize them to adapt to the images after resizing. We just need to divide the abscissa by the width and the ordinate at high.

3.3. Training Xception + FC

We use Adam optimizer to train the Xception + FC model with LR of 1e-4. And we uses MSE as loss function. After the final model converges, the loss on the verification set is reduced to 1.67e-4 with a training loss of 7.8104 e -5.

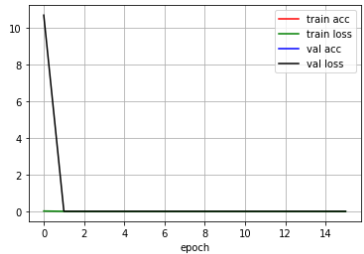

Figure 7. Loss curve of training process

In the experiment, we are surprised to find that the model's convergence speed of this simple method is very fast. Only 1 ~ 2 epoch can make this model had a smaller loss. But in order to fit the data better, we need to train more epochs. And it should be noted that the over-fitting of FC layer is very serious. Even with data augmentation. There is still a big gap between training set and validation set's loss.

3.4. Training U-net

Training U-net is relatively more complex. We need to generate heatmaps' labels in advance. From coordination labels to heatmap labels, the only we need is a sigma parameter of Gaussian distribution. There are some form of Gaussian function can be chosen.

$$g(x) = \frac{\alpha}{2\pi\sigma} Exp[-\frac{(x-x_0)^2+(y-y_0)^2}{2\sigma^2}]$$
$$g(x) = Exp[-\frac{(x-x_0)^2+(y-y_0)^2}{2\sigma^2}]$$

What we use in our experiment is (2), with specific sigma of 1.2.

Still, we use Adam as optimizer, and we use Binary Cross Entropy Loss as loss function. Because the heatmaps describes a probability distribution. Cross entropy can make the model converge better. Finally, the u-net model reach a loss of 1.6411e-4 in validation dataset, and 1.9105e-4 in train dataset.

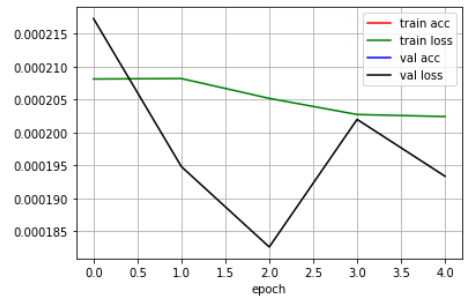

Figure 8. Loss curve of u-net training process (part)

We find that the heatmap regression method has better generalization ability. The loss on the validation set is less than that on the training set.

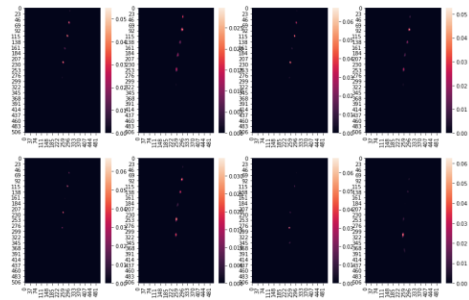

Figure 8. Heatmaps that u-net predicted

However, as described before, there are many Gaussian distribution points in the a predicted heatmap (Figure 8.) . These points will interfere with the selection of the position.

3.5. Combining Two Result

We know that multiplying any two probability distributions yields a probability distribution. When we have two distributions that describe the state of the system including noise, multiplying them will produce a distribution that can better reflect the state of the system. Therefore, we consider multiplying the result of Xception prediction with that of U-net prediction.

We assume that the coordinates predicted by Xception model are uncertain and obey a Gaussian distribution. Therefore, we transform the prediction of Xception into a heatmap of Gaussian distribution. The sigma parameter can be specify by experience or use a network to train them.

Then multiply the two heatmap to get a more accurate distribution heatmap (Figure 9. )

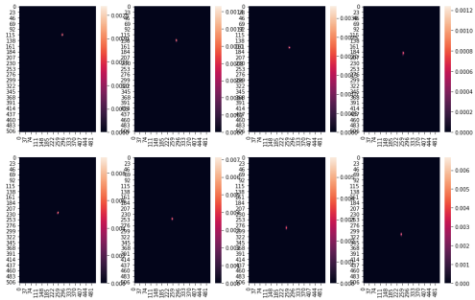

Figure 9. Heatmaps after multiplication

4. Estimation

We tested the Xception model on a test set with 50 images, each image have 11 landmarks. And it reached 71.3% precision (with 158 points' prediction bias > 8mm)

Then We combine Xception model with Heatmap Regression model, it shows a 99.6% precision on test images. In 550 landmarks, 2 landmarks are located incorrectly.

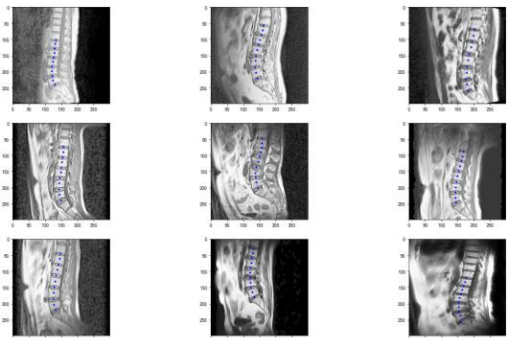

Figure 10. Landmark Localization Result

5. Conclusion

In the field of medical AI, the task of landmark localization take an essential role, but affected by the characteristics of medical images, it is challenging to locate landmarks in medical images. And medical AI often has higher requirements for the accuracy of landmark localization, otherwise the subsequent tasks will not be able to carry out.

Heatmap regression and direct coordinate regression are usually used in landmark localization tasks. The former has high positioning accuracy for Gaussian peak position, strong generalization of the model, but also has the disadvantages of high complexity, large consumption of time and space resources in the training process, and the post-processing is complex, because the heatmaps predicted usually has multiple Gaussian spots, which is easy to locate to the nearest point and cause large error. The latter is an end-to-end model with simple training and fast speed. After the model converges, the predicted points will be near the expected points, and the error with the key point will not be too large. However, it usually can not obtain high accuracy , and the FC layer aggravates over-fitting. It needs a lot of data to alleviate the over-fitting. And a deeper model is needed.

To improve two models accuracy. We can assume that the predicted results of CNN + FC model obey a Gaussian distribution. And generator a Gaussian heatmap for each predicted coordinate. Then multiply it with heatmap created by image-to-image mapping model (such as FCN, U-net) we can get a more accurate prediction. When the two probability distributions which describes the system states are multiplied, we can get a distribution which can better reflect the state of the system.